\documentclass{article} % For LaTeX2e
\usepackage{iclr2022_conference,times}

\usepackage{amsmath,amsfonts,bm}

\def\eqref#1{equation~\ref{#1}}
\def\1{\bm{1}}

\DeclareMathAlphabet{\mathsfit}{\encodingdefault}{\sfdefault}{m}{sl}
\SetMathAlphabet{\mathsfit}{bold}{\encodingdefault}{\sfdefault}{bx}{n}

\newcommand{\R}{\mathbb{R}}

\DeclareMathOperator*{\argmax}{arg\,max}
\DeclareMathOperator*{\argmin}{arg\,min}

\usepackage{mathtools}
\usepackage{booktabs}       % professional-quality tables
\usepackage{helvet} % DO NOT CHANGE THIS
\usepackage{courier}  % DO NOT CHANGE THIS
\usepackage[hyphens]{url}  % DO NOT CHANGE THIS
\usepackage{graphicx} % DO NOT CHANGE THIS
\usepackage{subfigure} % DO NOT CHANGE THIS
\usepackage{wrapfig} % DO NOT CHANGE THIS
\usepackage{hyperref}

\usepackage{algorithm}
\usepackage{amsmath}
\usepackage{comment}
\usepackage{adjustbox}
\usepackage{algorithmic}
\usepackage{amsfonts}       % blackboard math symbols
\usepackage{caption} % DO NOT CHANGE THIS AND DO NOT ADD ANY OPTIONS TO IT
\usepackage{bm}
\newcommand{\D}{\bm{\mathcal{D}}}
\newcommand{\Lmd}{\Lambda}

\usepackage{xcolor}
\usepackage{selectp}
\title{Improving Hyperparameter Optimization\\ by Planning Ahead}

\author{Hadi S. Jomaa, Jonas Falkner \& Lars Schmidt-Thieme \\
Department of Computer Science\\
University of Hildesheim\\
31141 Hildesheim, Germany \\
\texttt{\{hsjomaa,falkner,schmidt-thieme\}@ismll.de} \\
}

\begin{document}

\maketitle

%%% ABSTRACT %%%
\begin{abstract}
Hyperparameter optimization (HPO) is generally treated as a bi-level optimization problem that involves fitting a (probabilistic) surrogate model to a set of observed hyperparameter responses, e.g. validation loss, and consequently maximizing an acquisition function using a surrogate model to identify good hyperparameter candidates for evaluation. The choice of a surrogate and/or acquisition function can be further improved via knowledge transfer across related tasks. In this paper, we propose a novel transfer learning approach, defined within the context of model-based reinforcement learning, where we represent the surrogate as an ensemble of probabilistic models that allows trajectory sampling. We further propose a new variant of model predictive control which employs a simple look-ahead strategy as a policy that optimizes a sequence of actions, representing hyperparameter candidates to expedite HPO. Our experiments on three meta-datasets comparing to state-of-the-art HPO algorithms including a model-free reinforcement learning approach show that the proposed method can outperform all baselines by exploiting a simple planning-based policy. 
\end{abstract}

%%% #1 %%%
% x establishing an equivalence  
% x difference in MDPS
% x explicit planning
% x continuous vs discrete faster for transfer learning

% x ahat missing
% x pi used to select action
% learn the stqte vs specific engineering the state
\section{Introduction}
Hyperparameter optimization (HPO) is a ubiquitous problem within the research community and an integral aspect of tuning machine learning algorithms to ensure generalization beyond the training data. HPO is often posed as a sequential decision-making process, however it can be seen as a special use-case of model-based reinforcement learning (MbRL)~\citep{Sutton1991_Dyna,henaff2017model} developed under the guise of some idiosyncratic terms.
%\textcolor{blue}{For more than a decade~\citep{Hutter2011_SMAC,Bergstra2011_Algorithms}, hyperparameter optimization (HPO) has been framed as a sequential decision-making process, whereas to a large extent, the community has overlooked the fact that HPO is a specific use-case of model-based reinforcement learning (MbRL)~\citep{Sutton1991_Dyna,henaff2017model} that has been developed under the guise of some idiosyncratic terms.}

%\textcolor{blue}{
In MbRL the objective is to train a \textit{transition model} to approximate an underlying transition function via interactions with an environment governed by some \textit{policy}, e.g. random shooting~\citep{Nagabandi2018_NNDynamics}. During inference, an agent navigates the simulated environment to optimize a pre-defined \textit{reward function}, while the transition model remains unchanged. Conventionally in HPO, a \textit{surrogate model} is trained to estimate some black-box function, e.g. validation loss of a machine learning algorithm under investigation~\citep{Rasmussen2003_GP, Snoek2015_Scalable, Springenberg2016_Bohamiann}. An \textit{acquisition function}~\citep{Wilson2017_Marginal} interacts with the surrogate model to propose potential hyperparameters that optimize the black-box response, viewed as a \textit{reward function}. Effectively, the  surrogate model \textbf{is} the only unknown component for a transition model, that prevents HPO from being framed fully as MbRL problem.%}%Recently, an extensive study has pointed out the underlying empirical assumptions that are taken for granted without explicit motivation~\cite{cowen2020hebo}. 

 In this paper we present a novel formulation for HPO defined within the context of MbRL. Namely, we learn an ensemble of probabilistic neural network models~\citep{Lakshminarayanan2017_DE} and show that using model predictive control (MPC)~\citep{Kamthe2018_ModelPredictive} and a novel look-ahead variant to navigate the simulated black-box environment, we can outperform conventional Bayesian optimization techniques with heuristic acquisition functions in both transfer and non-transfer learning settings. 
 Thus, we elaborate on the importance of \textit{explicit} planning in HPO that has been largely overlooked by the community. We also formally define HPO as a Markov decision process (MDP) with a simple, yet novel, state representation as the \textit{set} of previously evaluated hyperparameters and their corresponding responses. 
 
\begin{figure}
    \centering
    \includegraphics[width=0.98\columnwidth]{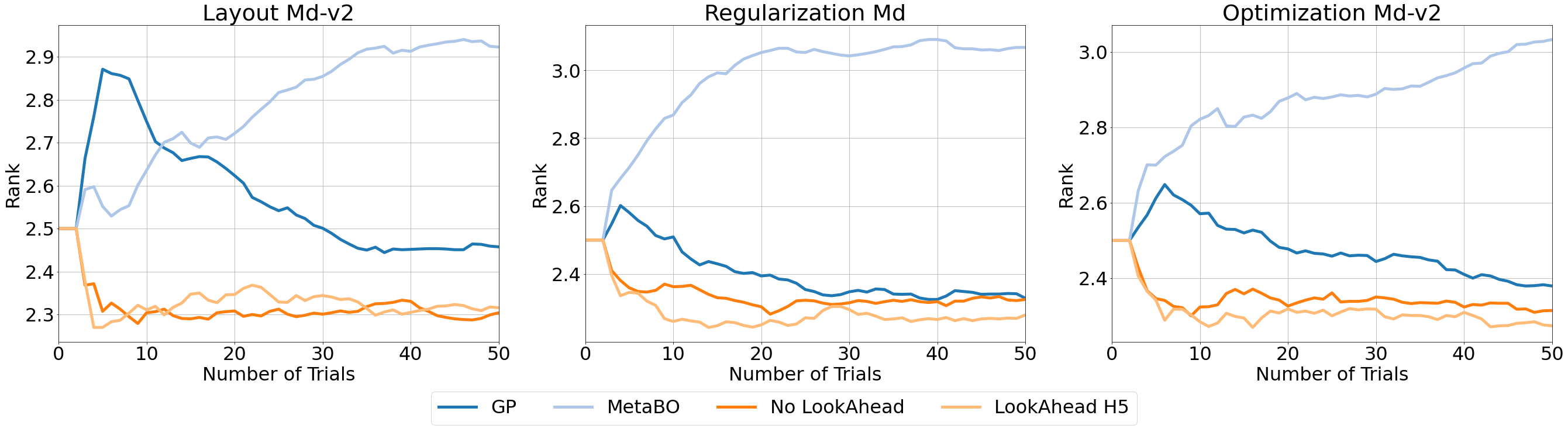}
    \caption{Motivating the effect of planning for HPO.}
    \label{fig:motivation}
\end{figure}

 We argue that with a clearly defined transition model, we can replace the acquisition function with a simple policy that maximizes the reward across the simulated trajectories, and achieve better results through proper planning, as shown in Figure~\ref{fig:motivation}. LookAhead MPC-5 is our look-ahead strategy with MPC that simulates 5 states ahead. We provide full experimental details in Section~\ref{sec:experiments} and here focus on the motivation.
 
\begin{comment}
  Recent attempts at modeling HPO as an MDP have approached the problem with model-free RL~\citep{Volpp2020_Metabo,Jomaa2019_HYPRL}, relying on an engineered state representation that is heavily influenced by an underlying surrogate model. However, these approaches are brittle and do not scale to large domains as shown by~\cite{Wistuba2021_FSBO,Jomaa2021_DMFBS}. Additionally, the existing literature on HPO defines the surrogate solely over the domain of hyperparameters and relies on engineered acquisition functions to select new candidates to evaluate. Each acquisition function exploits the statistics of the estimated response surface differently, leading to drastically different results, partly due to the nature of the surrogate itself. We argue that with a proper state transition model, we can replace the acquisition function with a simple policy that maximizes the reward across trajectories simulated by the model, and achieve better results through better proper planning, as shown in Figure~\ref{fig:motivation}. We provide full experimental details in Section~\ref{sec:experiments} and here focus on the motivation.
\end{comment}
\noindent
Our main contributions are summarized as i) a formal definition of HPO as an MDP which does not depend on any engineered heuristics,
  ii) a new transfer learning surrogate model represented by an ensemble of probabilistic neural networks,
  iii) a novel acquisition function that implements a look-ahead strategy paired with model predictive control,
  iv) clear motivation that highlights the impact of planning in HPO, which to the best of our knowledge, has never been addressed before.
 
%%% #2 %%%
\section{Related Work}
Hyperparameter optimization has been extensively studied in the community within two main settings: single task (non-transfer learning) HPO and transfer learning HPO. 

For single task HPO, solutions often involve estimating the hyperparameter response surface using a (probabilistic) surrogate, such as a  Gaussian process~\citep{Rasmussen2003_GP,Bergstra2011_Algorithms}, random forests~\citep{Hutter2011_SMAC}, Bayesian neural networks~\citep{Springenberg2016_Bohamiann}, or some hybrid approach~\citep{Snoek2015_DNGO}. Hyperparameter candidates are then selected via an acquisition function~\citep{Wilson2017_Acquisitions}, e.g. expected improvement, that satisfies some well-motivated assumptions and utilizes the statistics from the posterior to score each hyperparameter. 

Transfer learning solutions on the other hand leverage available data from experiments on related tasks to expedite HPO. Simple solutions involve initializing the surrogate using hyperparameters that perform well on datasets that have similar meta-features~\citep{Feurer2014_Using, Jomaa2021_D2V}, i.e. dataset statistics. The response surface can also be modeled jointly in a multi-task setting~\citep{Bardenet2013_SCOT,Yogatama2014_Efficient,Perrone2018_ABLR}. For example, \cite{Salinas2020_CTS} transform the hyperparameter response into a similar distribution and learn a shared Gaussian Copula. Another way to achieve transfer learning is through a weighted combination of the surrogates~\citep{Wistuba2016_TSTR, Feurer2018_RGPE}. More recently, it has been shown that meta-learning an initialization of the surrogate by training to estimate the response across the training tasks, and subsequently fine-tuning to the target task improves generalization and leads to better performance~\citep{Wistuba2021_FSBO,Jomaa2021_DMFBS}. Additionally one can learn a transferable acquisition function~\citep{Volpp2020_Metabo,Wistuba2018_TAFR}.  \cite{Volpp2020_Metabo} propose a transferable acquisition function, MetaBO, as a policy that is meta-trained on related tasks. As a model-free RL approach, MetaBO is trained by interacting with the environment and observes an engineered state representation that is heavily influenced by underlying surrogates. MetaBO however is very sensitive to the number of trials~\citep{Wistuba2021_FSBO}.

Model-based reinforcement learning aims to learn a model of the environment by estimating the next state given a state-action pair. It is applied with great success in areas such as robotics and video games. Existing methods similarly involve learning a (probabilistic) model of the dynamics~\citep{Chua2018_PETS,Nagabandi2018_NNDynamics,Ko2007_GPRL}. Moreover, several methods improve the policy in a Dyna style approach~\citep{Sutton1991_Dyna}, by leveraging the generated trajectories from the model to mitigate the distribution shift. 

%Contrary to standard model-based reinforcement learning solutions, the environment for HPO is unknown and can only be queried. However, a common assumption is that it is a partially observable environment such that a transition model can be learned to navigate the joint response surface of the training tasks and the environment model can be improved through fine-tuning on the new task.

%%% #3 %%%
%%%%%%%%%%%%%%%%%%%%%%%%%%%%%%%%%%%
\section{Preliminaries}
\textbf{Hyperparameter Optimization:}
Let $\D$ denote the space of all datasets and $\Lmd\in\mathbb{R}^N$ the space of hyperparameters associated with an unknown black-box function $\ell^{(D)}:\Lmd\rightarrow\mathbb{R}$ where $\ell^{(D)}$ represents the response $\ell^{(D)}(\lambda)$, e.g. validation loss,  of a certain model under investigation trained on a dataset $D\in\D$ with hyperparameters $\lambda\in\Lmd$. For example, we might be interested in a neural network model where $\Lmd\coloneqq\mathbb{R}^+\times\mathbb{R}_0^+\times\mathbb{N}$ represents the learning rate, dropout rate, and the number of layers, respectively. The objective of HPO is hence to find the optimal hyperparameter configuration $\lambda^*:=\arg \min_{\lambda\in\Lmd}\ell^{(D)}(\lambda)$ given a fixed budget of $T$ trials.

HPO is commonly treated as sequential decision-making process, where a surrogate model $\hat{\ell}:\Lmd\rightarrow\mathbb{R}$ is iteratively fit to the history $\mathcal{H}^{(D)}_t\coloneqq\{\lambda_i,\ell^{(D)}(\lambda_i)\}_{i=1}^t$ of evaluated hyperparameters and an acquisition function  $\hat{w}: (\Lmd\times\mathbb{R})^*\rightarrow\Lmd$ is used to select the next candidate which minimizes the expected hyperparameter response: 
\begin{equation}
    \argmin_{\hat{w}}\mathbb{E}_{D\sim\rho_{\D}}~\ell^{(D)}(\{\lambda_1,\dots,\lambda_t\}),
\end{equation}
% ahat fix
where $\ell^{(D)}(\{\lambda_1,\dots,\lambda_t\}) := \min_{i\in\{1,\dots,t\}}\ell^{(D)}\left(\lambda_i\right)$, $\lambda_i:= \hat{w}(\{(\lambda_j,\ell^{(D)}(\lambda_j)\}_{j=1}^{i-1})$ and $\rho_{\D}$ is some distribution over the available datasets. Among the variety of acquisition functions, the \textit{expected improvement} is widely adopted~\citep{Mockus_Acquisitions}. %\textcolor{blue}{
Although HPO algorithms can be applied on both continuous and discrete search spaces, here we focus on discrete spaces because they allow for faster training in transfer learning settings~\citep{Schilling2016_PoG,Jomaa2021_DMFBS,Wistuba2021_FSBO}.%}

\textbf{Model-based reinforcement learning:} 
%algorithms use interactions with the environment to learn a model of the transitions. The tasks are formulated as a discrete-time Markov decision process (MDP) defined by the tuple
MbRL uses interactions with the environment to learn a parameterized approximation of the underlying transition function which subsequently can be used for planning. The learning task can be formulated as a discrete-time Markov decision process (MDP) defined by the tuple 
$\langle\mathcal{S},\mathcal{A},\tau,r\rangle$
%with $\mathcal{S}$ state space, $\mathcal{A}$ as the action space, $\tau:\mathcal{S}\times\mathcal{A}\rightarrow\mathcal{S}$ as the unknown transition
%function, and $\mathcal{R}:\mathcal{S}\times\mathcal{A}\rightarrow\mathbb{R}$ as the reward function.
with state space $\mathcal{S}$, action space $\mathcal{A}$, unknown transition
function $\tau:\mathcal{S}\times\mathcal{A}\rightarrow\mathcal{S}$ and reward function $r:\mathcal{S}\times\mathcal{A}\rightarrow\mathbb{R}$.
%The  aim  of
The general approach in RL is to learn the optimal policy $\pi$ 
%, such that expected cumulative discounted reward $J(\pi)=\mathbb{E} \sum_{t=0}^T\gamma^t r(s_t,a_t)$ is maximized, where $\gamma\in[0,1]$ denotes the discount factor.
maximizing the expected cumulative discounted reward given by $
    J(\pi):=\mathbb{E} \left[\sum_{t=1}^T\gamma^t r(s_t,a_t)\mid s_t=\tau(s_{t-1},a_{t-1}),a\sim\pi(s_t)\right]
$ 
where $s_t\in\mathcal{S}$ and $a_t\in\mathcal{A}$  with discount factor $\gamma\in[0,1]$. We note that the explicit reward function or an \textit{approximate} reward model is necessary for planning actions under some learned transition model. 
%Denote by $\hat{\tau}$ the model which approximates the unknown transition function, i.e.\ the model of the environment, that generates an estimate of 
%the next state given action $a_t$ taken in state $s_t$. During inference, the sequence of actions to take is presented as solution of the following optimization problem:
%\begin{equation}
%    \arg\max_{a_1,\dots,a_T}\sum_{t=1}^{T}\mathcal{R}\left(\hat{\tau}(s_{t-1},a_{t-1}),a_t\right)
%\end{equation}
During inference, the optimal sequence of actions is commonly presented as the solution to the following optimization problem:
\begin{equation}
    \label{eq:standard-reward}
    \argmax_{a_1,\dots,a_T}\sum_{t=0}^{T-1}\hat{r}\left(
    \hat{\tau}(s_t,a_t),\ a_{t+1}
    \right),
\end{equation}
which leverages the transition model $\hat\tau$ to predict the next state $s_{t+1}$ given a state $s_t$ and action $a_t$.

%%% #4 %%%
%%%%%%%%%%%%%%%%%%%%%%%%%%%%%%%%%%%
\section{Hyperparameter optimization via Model-Based RL} 
%In this paper, we formulate HPO as an MDP, since its solutions are often designed as a sequential decision-making process that rely heavily on the observed hyperparameter response(s). Formally, we can define an observed state $s_t\in \mathcal{S}\subseteq\left(\Lmd\times\mathbb{R}\right)^*$ simply as the history of evaluated hyperparameters $\lambda$ at time $t$ and their corresponding responses $\ell(D,\lambda)$, s.t $s_t:= \{\left(D,\lambda_i,\ell(D,\lambda_i)\right)\}_{i=1:t}$ for a given dataset $D\in\D$. 
%We clearly formulate %hear <- haha I ss you are tired :P
In the following we clearly formulate
%here 
HPO as an MDP which in turn facilitates the use of a variety of RL approaches to solve the problem. %is a sequential decision-making process it can also be formulated 
As a sequential decision-making process, hyperparameters $\lambda\in\Lambda$ are iteratively selected and evaluated forming a simple MDP. Particularly, the state space is defined as ${\cal S} := \D\times(\Lambda\times\R)^*$ with \begin{equation}
    s_t:=\left(D,((\lambda_1,\ell_1),\dots,(\lambda_t,\ell_t))\right) 
\end{equation} where $\ell_i:=\ell^{(D)}(\lambda_i)$ and action space ${\cal A}  := \Lambda$.
The  ground truth transition function $\tau: {\cal S} \times \Lambda \rightarrow {\cal S}$ simply appends the new observations to the previous ones,
\begin{equation}
    \tau(s_t,\lambda):=\left(D,((\lambda_1,\ell_1),\dots,(\lambda_t,\ell_t),(\lambda,\ell^{(D)}(\lambda)))\right)
\end{equation} for dataset $D\in\D$ and a ground truth reward function, $r:=\mathcal{S}\times\Lambda\rightarrow\mathbb{R}$, that returns the loss reduction of the new hyperparameters over the best loss so far,
\begin{equation}
    r(s_t,\lambda):=\max \{0,\min_{i=1:t}\ell_i - \min\{\ell_1,\dots,\ell_t,\ell^{(D)}(\lambda)\}\}
\end{equation}

To estimate both, the transition function and the reward function, as done in MbRL, the only missing piece is a model for the validation loss, the so called surrogate model. Given $\hat{\ell}^{(D)}:\Lambda\rightarrow\mathbb{R}$, the transition and reward
models are just:  $\hat{\tau}: {\cal S} \times \Lambda\rightarrow {\cal S}$ where
\begin{equation}
    \hat{\tau}(s_t,\lambda):=\left(D,((\lambda_1,\ell_1),\dots,(\lambda_t,\ell_t),(\lambda,\hat\ell^{(D)}(\lambda)))\right)
\end{equation} and $\hat{r}:\mathcal{S}\times\Lambda\rightarrow\mathbb{R}$ with
\begin{equation}
    \hat{r}(s_t,\lambda):=\max \{0,\min_{i=1:t}\ell_i - \min\{\ell_1,\dots,\ell_t,\hat\ell^{(D)}(\lambda)\}\}.
\end{equation}
In this sense, HPO using a surrogate model can be seen as a special case of MbRL,
and thus all approaches researched for MbRL can be directly
applied to HPO. 

If there is no information about the dataset $D$ available other than the previously observed hyperparameter responses, then the surrogate can be formed as $\hat\ell:\Lambda\times(\Lambda\times\mathbb{R})^*\rightarrow\mathbb{R}$.

%As in any black-box optimization problem, the amount of information is usually restricted to the evaluated iterates and their observed responses. Thus, for simplicity we define the observation space $\mathcal{S}\subseteq\left(\Lmd\times\mathbb{R}\right)^*$ as the \textbf{set} of evaluated hyperparameters $\lambda$ up to time $t$ and their corresponding responses $\ell(\lambda, D)$ for a given dataset $D\in\D$. In other words, the state is given by the \textit{history} $s_t=\mathcal{H}_t$.
%, s.t $s_t:= \{\left(D,\lambda_i,\ell(D,\lambda_i)\right)\}_{i=1:t}$ Furthermore, we define the action space as the hyperparameter search space $\mathcal{A}\coloneqq\Lmd$, where an action $a_t\in \mathcal{A}$ corresponds to the hyperparameter to be evaluated next for the model under investigation.
%Upon the execution of the selected action $a_t$, we receive the reward $r(s_t,a_t)\in\mathcal{R}$ as a function of the hyperparameter response. Upon execution of the selected action $a_t$, the environment transitions to a new state $s_{t+1}=s_t\cup\{\left(a_t,\ell(a_t, D)\right)\}$ composed of the previous state and the newly selected hyperparameter and its observed response. Finally, we define the reward function in terms of the observed response, such that $r(s_t,a_t)\in\mathcal{R}$. The key idea is to select an action that minimizes the validation loss (or maximizes the validation accuracy). 

% train for GP specific insted of operating on the real state, the authors derive some engineered state based on heuristics hihglihgt assumptions
%\textcolor{blue}{
To the best of our knowledge, so far there exist two prior publications which attempt to solve HPO using RL by defining distinct MDPs. \cite{Jomaa2019_HYPRL} originally define the state as the dynamic history of evaluated hyperparameters and their respective responses as $\mathcal{S}:=\D\times(\Lambda\times\mathbb{R})^*$ and $s_t$ as defined earlier. However, they suggest that the order by which the hyperparameters are selected impacts the decision on which action to take next, which is why they model this temporal aspect via an LSTM in their policy. Their transition function generates new states by appending the action and the observed response to the previous state according to $\tau(s_t,\lambda):=\left(D,((\lambda_1,\ell_1),\dots,(\lambda_t,\ell_t),(\lambda,\ell^{(D)}(\lambda)))\right)$. \cite{Volpp2020_Metabo} represent the state space as $\mathcal{S}:=(\mathbb{R}^M)^*$ with $s_t:=((\mu_t(\lambda_i),\sigma_t^2(\lambda_i),\lambda_i,\psi_i))_{i=1}^{\mid\Lambda\mid}$ where $\mu_t$ and $\sigma_t^2$ are the mean and variance of the posterior distribution of an underlying surrogate and $\psi_i$ are some engineered attributes. The transition function simply updates the parameters of the surrogate based on the new observations. %$\tau$ After an action is selected, the surrogate is fine-tuned to the new hyperparameters, and consequently the attributes of the hyperparameters on the grid change. %The environment presented in this approach is heavily engineered, with a transition function that is dependent on an underlying surrogate selected under unmotivated assumptions~\citep{cowen2020hebo}. 
Both approaches define the action space as the discrete grid of hyperparameters, $\mathcal{A}:=\Lambda$, whereas the reward function is computed in terms of the regret.

In strong contrast to these methods that replace the standard acquisition function with a policy trained via model-free RL approaches, we design a %proper and 
powerful yet simple transition function for HPO and train a transition model that facilitates planning to improve HPO in the context of MbRL.
\section{MetaPETS Algorithm}
In this section, we present our model-based reinforcement learning algorithm. Namely, we define the transition function, our novel look-ahead acquisition function, and the associated training procedure.

%%%
\subsection{Probabilistic Transition Function}
%The objective of model-based reinforcement learning is to learn the dynamics of the environment, i.e the dynamics or transition function 
Model-based reinforcement learning is mainly concerned with learning a parameterized model of the environment which approximates the underlying dynamics, or transition function,
$\tau:\mathcal{S}\times\mathcal{A}\rightarrow\mathcal{S}$, 
such that the next state can be estimated given the current %a defined 
state-action pair. Following the standard RL notation, we denote by actions $a$ hyperparameters $\lambda$, i.e.\ $\ell^{(D)}(a)=\ell^{(D)}(\lambda)$. %As such 
Accordingly, the choice of the model class plays an important role. 
A common approach to model the environment is through Gaussian processes (GP)~\citep{Rasmussen2003_GPRL, Ko2007_GPRL, Boedecker2014_Approximate} or a mixture of GPs~\citep{Khansari2011_GPMix} 
which provide uncertainty estimates for the predictions and, more importantly, for unexplored areas. 
However, such models suffer from the curse of dimensionality and thus are difficult to scale to high-dimensional domains. 
Deep neural networks on the other hand have shown great success in handling uncertainty estimation~\citep{Springenberg2016_Bohamiann, Gal2016_Dropout, Lakshminarayanan2017_DE}, with success in predictive modeling of images~\citep{Watter2015_Embed} and short-horizon control tasks
for high-dimensional data.

\begin{wrapfigure}{r}{0.4\textwidth}
  \begin{center}
    \includegraphics[width=0.38\textwidth]{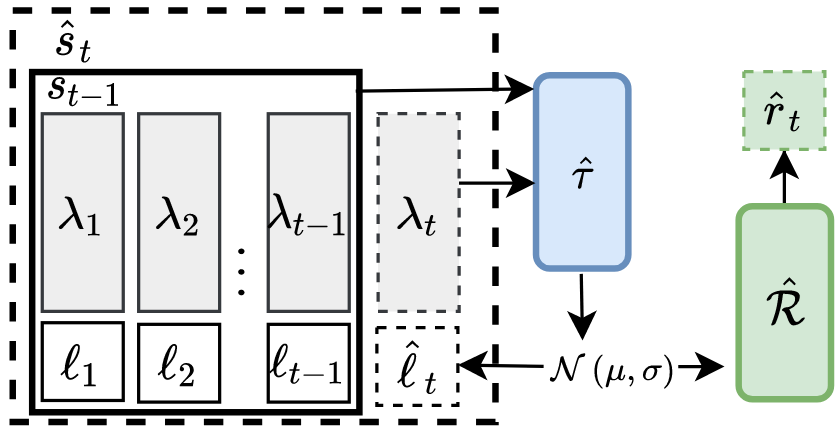}
  \end{center}
    \caption{Architecture of the transition function}
    \label{fig:transition}
\end{wrapfigure}

%To maintain 
%In order to increase training robustness, transition models are conventionally trained to estimate the difference between the current state and the next state 
%, s.t. $\hat{s}_{t+1}:=s_{t}+\hat{\tau}(s_t,a_t)$.  $\Delta \hat{s}_{t+1} \coloneqq \hat{s}_{t+1}-s_t$. 
%This particular trick is not feasible when the state is represented by a set. 
%Unfortunately, this approach is not feasible in our case where {\color{blue} we choose to represent the state as a set.}
In our case, the state is represented as the history of selected hyperparameters and their responses, up to a given time, while the actions consist of the available hyperparameters to be evaluated. Hence there is only one missing piece of information that is required to establish the next state, and that is the response of the respective selected action (Figure~\ref{fig:transition}). This missing element we acquire by training the transition model to directly estimate the response for a given state-action pair.
%Nevertheless, since the state is represented as the history of the selected hyperparameters and their responses whereas actions correspond to the next hyperparameter to be evaluated, the only missing component which the transition model has to provide is the unknown response $\ell(a, D)$ for each possible action $a \in \Lmd$, see Figure~\ref{fig:transition}.
%Finally, we define the reward $\mathcal{R}$ as the normalized regret, i.e. the distance to the global minimum, defined as :
%We parameterize our transition function as a \textit{probabilistic} neural network that outputs the components of a probability distribution function, usually a Gaussian distribution. 
Specifically, we parameterize our transition function   $\hat{\tau}_\theta(s_t,a_t):\mathcal{S}\times\mathcal{A}\rightarrow\mathbb{R}_0^+\times\mathbb{R}^+$ as a \textit{probabilistic} neural network with parameters $\theta$, which takes as input the current state-action pair $(s_t, a_t)$ and outputs the defining parameters of a probability distribution function, e.g.\ a Gaussian distribution. 
%The governing assumption is that the instance is derived from a heteroscedastic Gaussian distribution, i.e. the variance varies between instances. In this well-established approach, the parameterized neural network produces two outputs, namely the mean $\mu_\theta(x)$ and variance $\sigma_\theta(x)^2$ and is trained by minimizing the negative log-likelihood: 
The governing assumption for this well-established approach is that the instance is derived from a heteroscedastic Gaussian distribution, i.e.\ the variance is not fixed between instances. Accordingly, the network outputs the respective mean $\hat{\mu}_\theta(x)\in\mathbb{R}_0$ and variance $\hat{\sigma}^2_\theta(x)\in\mathbb{R}^+$ and is trained by minimizing the negative log-likelihood: 
\begin{equation}
    \mathbb{E}_{D\sim\rho_{\D}}  \left[-\frac{\log\hat{\sigma}_\theta(s,a)^2}{2} + \frac{(\ell^{(D)}(a)-\hat{\mu}_\theta(s,a))^2}{2\hat{\sigma}_\theta(s,a)^2} + \text{ct}\right].
\end{equation}
%Thus, the transition function is denoted as
Thus, we define the \textit{predicted} hyperparameter response as $\hat\ell(a)\sim\mathcal{N}\left(\hat{\mu}(\cdot,a),\hat{\sigma}^2(\cdot,a)\right)$. 

\textbf{The Case for Ensembles:} 
%{\color{blue} (needs to be rewritten/improved, i.e. making the case for better uncertainty estimation vs. ensemble just for better predictions. Also need to reference the standard PETS algorithm here)} 
%Probabilistic neural networks are good at capturing the aleatoric uncertainty in the data. However, training a single neural network for certain tasks introduces epistemic uncertainty which arises due to the inherently stochastic nature of the training process. A principled approach to deal with this issue is by bootstrapping models in an ensemble. An ensemble of deterministic models, such as neural networks, reduces bias and reduces epistemic uncertainty. However, the predictions of single models would not capture the aleatoric uncertainty. It has been shown that simply bootstrapping an ensemble of probabilistic neural networks outperforms standard probabilistic networks as well as other Bayesian-based approaches.
Probabilistic neural networks can capture and model the \textit{aleatoric} uncertainty of the environment (e.g.\ observation noise) by parameterizing a suitable distribution. However, a single probabilistic network is not capable of capturing any \textit{epistemic} uncertainties (e.g.\ lack of a sufficient amount of data to completely determine the real dynamics). In particular, the epistemic uncertainty vanishes in the limit of infinite data but can have a major impact in low data regimes like HPO. For that reason, it is of major importance to account for epistemic uncertainty. 

A principled approach to deal with this issue is by bootstrapping models in an ensemble. An ensemble of deterministic models, such as standard neural networks, reduces bias and epistemic uncertainty. 
Therefore, to capture both, aleatoric as well as epistemic uncertainty, we employ a simple bootstrapped ensemble of probabilistic neural networks, which has been shown to outperform single probabilistic networks as well as other Bayesian approaches~\citep{Chua2018_PETS}.

Given such an ensemble of probabilistic transition models denoted as $\{\tau_{\theta_i}\}_{i=1}^{N_E}$, we can observe a mixture of $N_E$ distributions, i.e. $\{\mathcal{N}\left(\mu_{\theta_i},\sigma_{\theta_i}^2\right)\}_{i=1}^{N_E}$. Following \cite{Lakshminarayanan2017_DE} we aggregate the outputs as
    \begin{equation}
        \mu_*(\cdot,a) = \frac{1}{N_E}\sum_{i=1}^{N_E}\mu_{\theta_i}(\cdot,a)
    \end{equation} and 
    \begin{equation}
        \sigma^2_*(\cdot,a) = \frac{1}{N_E}\sum_{i=1}^{N_E}\left(\sigma_{\theta_i}^2(\cdot,a) + \mu_{\theta_i}^2(\cdot,a)\right) - \mu_*^2(\cdot,a)
    \end{equation} leading to an \textit{predicted} hyperparameter response in form of $\hat\ell(a)\sim\mathcal{N}\left(\mu_*(\cdot,a),\sigma^2_*(\cdot,a)\right)$. We want to point out that the explicit advantage of the ensemble approach in particular is the simple but excellent estimation of uncertainty which constitutes one of the main performance drivers in HPO in recent years.

\subsection{Planning with the Learned Transition Function}
%Given a model of the environment that facilitates the creation of imaginary (simulated) experiences, the second aspect of model-based RL is to design and improve a policy that exploits such a model. 
Given the learned dynamics model, we subsequently require an approach that can effectively leverage the acquired information to predict the optimal sequence of hyperparameters to evaluate.
For example, Dyna-style algorithms~\citep{Sutton1991_Planning} use the simulated experience to mitigate the distribution shift~\citep{Luo2019_Algorithmic, Kurutach2018_ModelEnsemble, Clavera2018_ModelBased} by integrating trajectories that have been generated via the policy to further train the transition model. 

%For simplicity, we use model-predictive control, that is given the transition function, which simultaneously provides us with the expected response as well as help generate the next state,  we use random shooting~\citep{Zhou2014_RandomShooting} to solve the following optimization objective: 
\subsubsection{Model Predictive Control}
For simplicity, here we consider the well-known approach of \textit{model-predictive control (MPC)} which has been used in many complex control scenarios \citep{Bouffard2012_Learning, Lenz2015_DeepMPC, Amos2018_DifferentiableMPC}. Furthermore, it is easy to implement and does not require any gradient computation. 

The general idea of MPC is to solve an optimization problem in a specific horizon on top of the learned transition model to produce a sequence of actions.
In particular we employ a simple \textit{random shooting (RS)}~\citep{Zhou2014_RandomShooting} technique to solve the following optimization objective: 
\begin{equation}
    \argmax_{a_1,\dots,a_T}\hat{r}(s_{T},a_{T},D).
\end{equation}
Notice that since the reward is measured in terms of the regret %(Equation~\ref{eq:reward}), 
we do not need to sum over the rewards at each simulated state, as presented in Equation~\ref{eq:standard-reward}, but simply observe the reward at the final state in the rollout.

%The policy finds the optimal trajectory that maximizes the reward, but in practice, we select the first action of the sequence and recalculate the trajectory at the next state. 
The RS policy generates $K$ random action sequences with a horizon of length $H$ from a uniform distribution and evaluates them via the learned transition function. Then usually only the first action of the best candidate sequence is executed and the procedure is repeated from the new state. Finally, we also fine-tune the transition model(s) given the newly evaluated hyperparameters on the test datasets to mitigate the distribution shift, which is common practice in all Bayesian-based optimization techniques.
%Given the ensemble, we can compute out state as the average across the models, i.e. 

\subsubsection{LookAhead MPC}
Conventional acquisition functions are agnostic to the order by which the previous hyperparameters \begin{wrapfigure}{r}{0.3\textwidth}
  \begin{center}
    \includegraphics[width=0.28\textwidth]{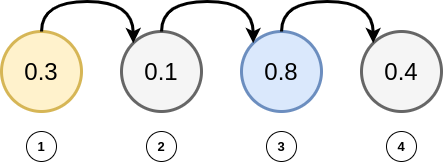}
  \end{center}
  \caption{Simulated Trajectory rewards. MPC selects the first item. LookAhead MPC selects the third item.}
  \label{fig:lookahead}
\end{wrapfigure} have been selected, i.e. $\hat{a}(H_t) = \hat{a}(\phi(H_t))$, where $\phi$ is some permutation function, since they are only concerned with the immediate improvement and do not account for how the optimization process evolves. MPC on the other hand normally selects the first action from the trajectory that had the highest reward at the end of the rollout. This can be slightly misleading as it implies that to arrive at the best hyperparameter configuration it is necessary to sequentially evaluate bad %\textit{worst} 
configurations. However, this is not true since the selection of hyperparameter configurations is independent and all configurations can be selected at all times. Thus, we propose to directly select the action that provides the highest estimated reward, i.e. lowest regret, from all the actions observed across all the simulated trajectories (Figure~\ref{fig:lookahead}).

\subsubsection{Trajectory Sampling}
The learned transition model outputs a distribution over the next state $s_{t+1}$. To leverage the uncertainty associated with the output response, we create particles by randomly sampling $p$ variations  as $s_{t+1}^p\sim\hat{\tau}(s_t,a_t)$ and propagate through the trajectory of every particle. At each step, the reward is then estimated as the average of the rewards across all \textit{particles}:
\begin{equation}
    \hat{r}(s_{t},a_{t}) = \frac{1}{P} \sum_{p=1}^P\hat{r}(s_{t}^{p},a_{t}).
\end{equation}

%%%
%\subsection{Training the Transition Function [In Progress!]}
\subsection{Training the Dynamics Model}
%As with other model-based reinforcement learning approaches, the first step in training the transition model is by generating random trajectories. We define the state $s_t$ as the collection of previously evaluated hyperparameters and their corresponding responses and use a deep set~\citep{Zaheer2017_DeepSets} formulation to encode the state into a fixed-size vector representation. 
The first step in our MbRL approach is to generate a dataset of transitions as training dataset for our dynamics model. We denote by $\mathcal{E}:=\{\left(D_n,\{\lambda_i,\ell^{(D_n)}(\lambda_i)\}_{i=1}^{T_{D_n}}\right)\}_{n=1}^N$ a meta-dataset of primary datasets, hyperparameters and their responses sampled from some unknown distribution of datasets $\rho_D$ and some unknown distribution of hyperparameters $\rho_{\Lmd}$. We generate our training dataset $\mathcal{E}^\text{train}$ from the meta-dataset by sampling a quadruple $(D_n,s_t,a_t, \ell^{(D_n)}(a_t))$ of datasets $D_n$, sets of hyperparameters and their responses as states $s_t$, unobserved hyperparameters as actions $a_t$, and the missing element from the next state $s_{t+1}$ as the response to $a_t$, i.e. $\ell^{(D_n)}(a_t)$. 
This is similar to the usually used approach of implementing a random policy to generate the trajectories~\citep{Nagabandi2018_NNDynamics}. Although it is possible to augment the $\mathcal{E}^\text{train}$ with trajectories generated using MPC, we noticed that it is \textbf{i)} time-consuming and \textbf{ii)} does not lead to a significant improvement compared to using simple random trajectories. We conjecture that this is due to the nature of our transition function. Specifically, because the sequence of hyperparameters selected (by any policy) does not immediately affect the output of the transition model, i.e. $\tau(s_{t-1},a) = \tau(s_{t},a)~\forall a\in\mathcal{A}$ but rather the output is only affected by the selected action.

Since our state $s_t$ is defined as the collection of previously evaluated hyperparameters and their corresponding responses, we employ a \textit{deep set}~\citep{Zaheer2017_DeepSets} formulation to encode $s_t$ into a fixed-size vector representation.
%Contrary to the conventional model-based RL algorithms, that predict the difference in the state, we estimate the response to a given action to facilitate the transition to the next state.
%In that regard, the state can be considered as context information~\citep{Kim2019_ANP}. Concretely, the transition model is defined as :
In contrast to the common approach of conventional model-based RL algorithms which are concerned with predicting the state difference $\Delta \hat{s}_{t+1}$, we estimate the response $\ell^{(D)}(a)$ to a given action $a$ while considering the state as context information~\citep{Kim2019_ANP}. 
Concretely, the transition model is defined as :
\begin{equation}
    \hat{\mu},\hat{\sigma}^2 = f\left(\left[a_t, \frac{1}{t}\sum_{i}^tg\left([\lambda_i,\ell_i]\right)\right]\right),
\end{equation}
such that $\hat{\tau}:=f\circ g$ where $g:=\Lmd\times\mathbb{R}\rightarrow\mathbb{R}^{N_g}$ and $f:\Lmd\times\mathbb{R}^{N_g}\rightarrow\mathbb{R}^+_0\times\mathbb{R}^+$ are feed-forward neural network models, and $[~]$ represents standard concatenation.
Given the variety of primary datasets, we use first-order meta-learning~\citep{Nichol2018_Reptile} to optimize the parameters of our transition model. We summarize the training procedure in Algorithm~\ref{alg:pseudocode}. %
 %%
%{\color{blue} (the paragraph below is not clear to me. Need to clarify and rewrite)}\\
%For standard Bayesian optimization techniques, the choice of the hyperparameters taken at a certain time affects the surrogate and indirectly also the maximizer of the acquisition function. The order in which the hyperparameters are selected up to a certain time $t-1$ does not affect the hyperparameter selected by the acquisition function at time $t$. 

%%% #6 %%%
%%%%%%%%%%%%%%%%%%%%%%%%%%%%%%%%%%%
\section{Experiments}
\label{sec:experiments}
The experiments are designed to address the following questions:
\begin{enumerate}
%    \item How well can a deep ensemble probabilistic network estimate the transition model? 
    \item Does the learned transition model generalize to new unseen tasks? 
%    \item Can we improve the learned transition model via adaptation to newly observed trajectory samples? 
    \item What is the importance of planning when doing HPO?
\end{enumerate}

\subsection{Meta-dataset}
%A meta-dataset is a collection of hyperparameters typically defined on a \textit{discretized} grid~\citep{Schilling2016_PoG,Jomaa2019_HYPRL} associated with a model under investigation, which has been evaluated offline by training the model with the mentioned hyperparameters on numerous \textit{primary} datasets and reporting an evaluation metric, e.g. validation loss. 
We evaluate our approach on three hyperparameter search spaces for feed-forward neural networks \citep{Jomaa2021_DMFBS}, that includes 120 UCI classification datasets~\citep{Asuncion2007_UCI}. We refer to the meta-datasets as \textbf{Layout Md-v2}, \textbf{Regularization Md}, and \textbf{Optimization Md-v2} which include the Cartesian product of the individual hyperparameters with a total of 324, 288, and 432 unique configurations, and 10, 7, and 12-dimensional hyperparameters, respectively. More information is presented in Appendix~\ref{appendix:metadataset}. %Figure~\ref{fig:kde} highlights the significant distribution shift of selected datasets across the meta-datasets. Additionally, the preprocessed domain of each meta-dataset includes 10, 7, and 12 dimensions respectively.

\subsection{Training the transition model}
%In Algorithm~\ref{alg:pseudocode}, we presented the pseudo-code for learning the transition model via first-order meta-learning. 
The transition model is trained via first-order meta-learning. Each meta-dataset is divided into 5 splits with 80 training datasets, 16 for validation and 24 for testing. We used a task batch size of 8 and 64 mini-batches per task. The number of inner iterations was set to 5. We use the ADAM~\citep{Kingma2014_ADAM} optimizer with a learning rate of $0.001$. The hyperparameters were tuned on the validation set and trained for $10000$ outer iterations with early stopping. We trained a total of $5$ distinctly initialized models for the ensemble using Tensorflow~\citep{Abadi2016_Tensorflow}. 
\begin{comment}

We report the average training and validation performance on the meta-datasets in Table~\ref{tab:losses} across the different splits. The standard deviation is computed across the different models.
\begin{table}[h]
\centering
\begin{adjustbox}{angle=0,scale=0.9} 
    \centering
    \begin{tabular}{lcc}
    \toprule
    Meta-Datasets &       NLL   &       MSE \\
    \midrule
    Layout Md-v2 (train) &  ~0.017 $\pm$ 0.006 &  0.050  $\pm$  0.001 \\
    Layout Md-v2 (valid) &  ~0.042 $\pm$ 0.005 &  0.055  $\pm$  0.001 \\
    \midrule
    \midrule
    Regularization Md (train) &  ~0.024 $\pm$ 0.004 &  0.051  $\pm$  0.001 \\
    Regularization Md (valid)&  ~0.053 $\pm$ 0.004 &  0.057  $\pm$  0.001 \\
    \midrule
    \midrule
    Optimization Md-v2 (train) & -0.011 $\pm$ 0.001 &  0.044  $\pm$  0.000 \\
    Optimization Md-v2 (valid) &  ~0.034 $\pm$ 0.002 &  0.053  $\pm$  0.000 \\    
    \bottomrule
    \end{tabular}
\end{adjustbox}
\caption{Training and validation performance}
\label{tab:losses}
\end{table}
\end{comment}

%\subsection{Hyperparameter Optimization}
\subsection{Baseline Models}
We compare against several single tasks and transfer learning baselines for HPO designed for black-box function optimization that rely only on the hyperparameters and the corresponding response. As such, some methods were not considered, e.g. \cite{Jomaa2021_DMFBS,Falkner2018_BOHB}. Particularly, we compare against, Random sampling~\citep{Bergstra2012_Random},
    GP~\citep{Rasmussen2003_GP},
    SMAC~\citep{Hutter2011_SMAC},
    BOHAMIANN~\citep{Springenberg2016_Bohamiann},
    TST-R~\citep{Wistuba2016_TSTR},
    %RGPE~\citep{Feurer2018_RGPE},
    ABLR~\citep{Perrone2018_ABLR},
    CTS~\citep{Salinas2020_CTS}, and 
    FSBO~\citep{Wistuba2021_FSBO}. A detailed overview can be found in Appendix~\ref{appendix:baselines}.

 We denote by LookAhead MPC-H\textbf{X}, our approach, which employs LookAhead MPC by navigating the simulated environment using random shooting up to the defined horizon \textbf{X} and then selects the first action of the trajectory that achieves the highest reward. We sample 1000 trajectories for our method. The models of the ensemble are iteratively fine-tuned to the new hyperparameters as all baseline models.% We also report the vanilla results of MPC-H\textbf{X}, the variant without any fine-tuning. 
\subsection{Evaluation Metrics}
\label{sec:eval}
We report two performance metrics to evaluate the effectiveness of the different baselines: \textbf{normalized regret} that represents the distance between the response of the evaluated hyperparameters and the optimal performance for each dataset, and the \textbf{rank} that measures the relative performance of each method compared to the rest of the baselines at each trial. The rank is computed at the task level and is agnostic to the heterogeneous ranges of the target response surfaces, thus can better signify the difference in performance.
\begin{comment}
\begin{enumerate}
    \item \textbf{Normalized Regret}: represents the distance between the response of the evaluated hyperparameters and the optimal performance for each dataset. Since we are using meta-datasets evaluated offline, the optimal performance is known. Formally, for a given trial $t$, the normalized regret is computed as:
    \begin{equation}
        \min_{\lambda\in\{\lambda_i\}_{i=1:t}} \frac{\ell(\lambda,D) - \ell_\text{min}(D)}{\ell_\text{max}(D) - \ell_\text{min}(D)}
    \end{equation}
    for a given dataset $D\in\D$. Since the response surface is normalized between $(0,1)$, the normalized regret at trial $t$ is simply $\min_{\lambda\in\{\lambda_i\}_{i=1:t}}\ell(\lambda,D)$. However, because this averages losses of different scales, few tasks might dominate, leading to deceiving results.
    
    \item \textbf{Rank}: measures the relative performance of each method compared to the rest of the baselines at each trial. The rank is computed at the task level and is agnostic to the heterogeneous ranges of the target response surfaces, thus can better signify the difference in performance.
\end{enumerate}
\end{comment}
\subsection{Results and Discussion}

\begin{table*}[]
    \centering
    \begin{adjustbox}{angle=0,scale=0.65} 

\begin{tabular}{lccccccccc}
\toprule
%%%%%%%%%%%%% rank computed withOUT rgpe
Methods &  \multicolumn{3}{c}{Layout Md-v2} &    \multicolumn{3}{c}{Regularization Md} &    \multicolumn{3}{c}{Optimization Md-v2} \\
{} &    15 &    33 &    50&15 &    33 &    50&15 &    33 &    50\\
\midrule
RS              &  6.48 $\pm$ \footnotesize{0.62}&  6.81 $\pm$ \footnotesize{0.38}&  6.78 $\pm$ \footnotesize{0.17} & 7.11 $\pm$ \footnotesize{0.49} &  7.33 $\pm$ \footnotesize{0.30}&  7.23 $\pm$ \footnotesize{0.21} & 7.19 $\pm$ \footnotesize{0.55}&  6.93 $\pm$ \footnotesize{0.44}&  7.07 $\pm$ \footnotesize{0.27}\\
BOHAMIANN       &  6.67 $\pm$ \footnotesize{0.05}&  6.48 $\pm$ \footnotesize{0.26}&  6.29 $\pm$ \footnotesize{0.17} & 6.51 $\pm$ \footnotesize{0.22} &  6.23 $\pm$ \footnotesize{0.14}&  5.86 $\pm$ \footnotesize{0.16} & 6.87 $\pm$ \footnotesize{0.35}&  6.55 $\pm$ \footnotesize{0.26}&  6.30 $\pm$ \footnotesize{0.17}\\
GP              &  6.28 $\pm$ \footnotesize{0.50}&  5.73 $\pm$ \footnotesize{0.19}&  5.72 $\pm$ \footnotesize{0.07} & 5.48 $\pm$ \footnotesize{0.19} &  5.32 $\pm$ \footnotesize{0.19}&  5.45 $\pm$ \footnotesize{0.16} & 5.92 $\pm$ \footnotesize{0.10}&  5.76 $\pm$ \footnotesize{0.11}&  5.53 $\pm$ \footnotesize{0.16}\\
SMAC            &  6.06 $\pm$ \footnotesize{0.24}&  6.19 $\pm$ \footnotesize{0.32}&  6.36 $\pm$ \footnotesize{0.18} & 6.22 $\pm$ \footnotesize{0.30} &  6.34 $\pm$ \footnotesize{0.06}&  6.43 $\pm$ \footnotesize{0.19} & 6.10 $\pm$ \footnotesize{0.26}&  6.24 $\pm$ \footnotesize{0.28}&  6.27 $\pm$ \footnotesize{0.29}\\
CTS             &  5.66 $\pm$ \footnotesize{0.25}&  5.74 $\pm$ \footnotesize{0.34}&  5.89 $\pm$ \footnotesize{0.22} & 5.80 $\pm$ \footnotesize{0.30} &  5.80 $\pm$ \footnotesize{0.15}&  5.98 $\pm$ \footnotesize{0.16} & 5.73 $\pm$ \footnotesize{0.26}&  6.16 $\pm$ \footnotesize{0.28}&  6.18 $\pm$ \footnotesize{0.27}\\
ABLR            &  7.11 $\pm$ \footnotesize{0.50}&  6.78 $\pm$ \footnotesize{0.36}&  6.55 $\pm$ \footnotesize{0.27} & 6.72 $\pm$ \footnotesize{0.39} &  6.36 $\pm$ \footnotesize{0.43}&  6.03 $\pm$ \footnotesize{0.21} & 5.80 $\pm$ \footnotesize{0.21}&  5.79 $\pm$ \footnotesize{0.20}&  5.72 $\pm$ \footnotesize{0.25}\\
TST-R           &  5.54 $\pm$ \footnotesize{0.23}&  5.42 $\pm$ \footnotesize{0.17}&  5.49 $\pm$ \footnotesize{0.13} & 5.50 $\pm$ \footnotesize{0.17} &  5.48 $\pm$ \footnotesize{0.09}&  5.48 $\pm$ \footnotesize{0.13} & 5.75 $\pm$ \footnotesize{0.20}&  5.59 $\pm$ \footnotesize{0.15}&  5.57 $\pm$ \footnotesize{0.13}\\
MetaBO          &  6.38 $\pm$ \footnotesize{0.26}&  6.85 $\pm$ \footnotesize{0.12}&  7.02 $\pm$ \footnotesize{0.15} & 7.02 $\pm$ \footnotesize{0.13} &  7.42 $\pm$ \footnotesize{0.15}&  7.60 $\pm$ \footnotesize{0.09} & 6.69 $\pm$ \footnotesize{0.15}&  7.00 $\pm$ \footnotesize{0.08}&  7.36 $\pm$ \footnotesize{0.08}\\
FSBO            &  \underline{5.31} $\pm$ \footnotesize{0.29}&  5.35 $\pm$ \footnotesize{0.27}&  \underline{5.31} $\pm$ \footnotesize{0.15} & 5.50 $\pm$ \footnotesize{0.14} &  5.45 $\pm$ \footnotesize{0.14}&  5.56 $\pm$ \footnotesize{0.17} & \underline{5.33} $\pm$ \footnotesize{0.17}&  5.39 $\pm$ \footnotesize{0.18}&  5.38 $\pm$ \footnotesize{0.28}\\
\midrule
LookAhead MPC-3 &  \textbf{5.18} $\pm$ \footnotesize{0.16}&  \textbf{5.31} $\pm$ \footnotesize{0.21}&  \textbf{5.26} $\pm$ \footnotesize{0.08} & \underline{5.10} $\pm$ \footnotesize{0.15} &  \textbf{5.13} $\pm$ \footnotesize{0.12}&  \textbf{5.15} $\pm$ \footnotesize{0.10} & 5.35 $\pm$ \footnotesize{0.15}&  \underline{5.36} $\pm$ \footnotesize{0.22}&  \underline{5.36} $\pm$ \footnotesize{0.24}\\
LookAhead MPC-5 &  5.32 $\pm$ \footnotesize{0.20}&  \underline{5.35} $\pm$ \footnotesize{0.08}&  5.32 $\pm$ \footnotesize{0.15} & \textbf{5.03} $\pm$ \footnotesize{0.14} &  \underline{5.14} $\pm$ \footnotesize{0.22}&  \underline{5.23} $\pm$ \footnotesize{0.11} & \textbf{5.27} $\pm$ \footnotesize{0.15}&  \textbf{5.23} $\pm$ \footnotesize{0.27}&  \textbf{5.26} $\pm$ \footnotesize{0.23}\\
\bottomrule
\end{tabular}
\end{adjustbox}

    \caption{Average rank. We report the best results in \textbf{bold} and \underline{underline} the second best.}
    \label{tab:rank}
\end{table*}
\begin{table*}[]
    \centering
    \begin{adjustbox}{angle=0,scale=0.65} 

\begin{tabular}{lccccccccc}
\toprule
Methods &  \multicolumn{3}{c}{Layout Md-v2} &    \multicolumn{3}{c}{Regularization Md} &    \multicolumn{3}{c}{Optimization Md-v2} \\
{} &    15 &    33 &    50&15 &    33 &    50&15 &    33 &    50\\
\midrule
RS             &  7.90 $\pm$ \footnotesize{1.59}&  5.59 $\pm$ \footnotesize{0.98} &  4.25 $\pm$ \footnotesize{0.64} &  8.54 $\pm$ \footnotesize{0.84} &  6.18 $\pm$ \footnotesize{0.56}&  4.51 $\pm$ \footnotesize{0.30} & 8.53 $\pm$ \footnotesize{0.89}&  5.40 $\pm$ \footnotesize{0.39}&  4.64 $\pm$ \footnotesize{0.32}\\
BOHAMIANN      &  8.05 $\pm$ \footnotesize{0.39}&  4.78 $\pm$ \footnotesize{0.17} &  3.55 $\pm$ \footnotesize{0.22} &  7.59 $\pm$ \footnotesize{0.39} &  4.30 $\pm$ \footnotesize{0.28}&  2.64 $\pm$ \footnotesize{0.40} & 8.64 $\pm$ \footnotesize{0.90}&  5.19 $\pm$ \footnotesize{0.48}&  3.75 $\pm$ \footnotesize{0.16}\\
GP             &  7.90 $\pm$ \footnotesize{1.06}&  4.25 $\pm$ \footnotesize{0.26} &  3.12 $\pm$ \footnotesize{0.18} &  6.05 $\pm$ \footnotesize{0.27} &  3.46 $\pm$ \footnotesize{0.30}&  2.49 $\pm$ \footnotesize{0.19} & 6.95 $\pm$ \footnotesize{0.37}&  4.51 $\pm$ \footnotesize{0.17}&  3.38 $\pm$ \footnotesize{0.20}\\
SMAC           &  7.56 $\pm$ \footnotesize{0.36}&  4.68 $\pm$ \footnotesize{0.47} &  3.66 $\pm$ \footnotesize{0.28} &  7.31 $\pm$ \footnotesize{0.77} &  4.81 $\pm$ \footnotesize{0.38}&  3.73 $\pm$ \footnotesize{0.44} & 6.96 $\pm$ \footnotesize{0.42}&  4.61 $\pm$ \footnotesize{0.33}&  3.69 $\pm$ \footnotesize{0.43}\\
CTS            &  6.39 $\pm$ \footnotesize{0.54}&  4.05 $\pm$ \footnotesize{0.64} &  3.32 $\pm$ \footnotesize{0.38} &  6.56 $\pm$ \footnotesize{0.59} &  3.82 $\pm$ \footnotesize{0.24}&  3.00 $\pm$ \footnotesize{0.29} & 6.67 $\pm$ \footnotesize{0.26}&  4.59 $\pm$ \footnotesize{0.39}&  3.58 $\pm$ \footnotesize{0.22}\\
ABLR           &  9.01 $\pm$ \footnotesize{0.88}&  5.60 $\pm$ \footnotesize{0.75} &  4.26 $\pm$ \footnotesize{0.44} &  7.99 $\pm$ \footnotesize{0.57} &  4.98 $\pm$ \footnotesize{0.52}&  3.53 $\pm$ \footnotesize{0.33} & 7.30 $\pm$ \footnotesize{0.33}&  4.68 $\pm$ \footnotesize{0.56}&  3.64 $\pm$ \footnotesize{0.55}\\
TST-R          &  6.70 $\pm$ \footnotesize{0.36}&  3.90 $\pm$ \footnotesize{0.36} &  2.91 $\pm$ \footnotesize{0.33} &  6.51 $\pm$ \footnotesize{0.39} &  3.88 $\pm$ \footnotesize{0.29}&  2.73 $\pm$ \footnotesize{0.39} & 6.78 $\pm$ \footnotesize{0.21}&  4.13 $\pm$ \footnotesize{0.23}&  3.21 $\pm$ \footnotesize{0.10}\\
MetaBO         &  7.24 $\pm$ \footnotesize{0.37}&  4.72 $\pm$ \footnotesize{0.32} &  3.86 $\pm$ \footnotesize{0.28} &  7.71 $\pm$ \footnotesize{0.11} &  5.72 $\pm$ \footnotesize{0.20}&  5.01 $\pm$ \footnotesize{0.21} & 8.63 $\pm$ \footnotesize{0.41}&  5.83 $\pm$ \footnotesize{0.24}&  5.13 $\pm$ \footnotesize{0.19}\\
%RGPE           &  \textbf{5.64} $\pm$ \footnotesize{0.80}&  \textbf{3.51} $\pm$ \footnotesize{0.26} &  \underline{2.63} $\pm$ \footnotesize{0.40} &  \textbf{5.37} $\pm$ \footnotesize{0.51} &  \textbf{2.79} $\pm$ \footnotesize{0.17}&  \textbf{1.79} $\pm$ \footnotesize{0.08} & \underline{6.37} $\pm$ \footnotesize{0.47}&  4.16 $\pm$ \footnotesize{0.46}&  3.24 $\pm$ \footnotesize{0.19}\\
FSBO           &  6.41 $\pm$ \footnotesize{0.50}&  \textbf{3.60} $\pm$ \footnotesize{0.31} &  \textbf{2.48} $\pm$ \footnotesize{0.29} &  5.99 $\pm$ \footnotesize{0.57} &  3.18 $\pm$ \footnotesize{0.14}&  2.32 $\pm$ \footnotesize{0.26} & \textbf{6.06} $\pm$ \footnotesize{0.37}&  \underline{3.90} $\pm$ \footnotesize{0.30}&  \underline{3.04} $\pm$ \footnotesize{0.41}\\
\midrule
LookAhead MPC-3           &  \textbf{6.15} $\pm$ \footnotesize{0.49}&  \underline{3.65} $\pm$ \footnotesize{0.42} &  \underline{2.60} $\pm$ \footnotesize{0.32} &  \underline{5.50} $\pm$ \footnotesize{0.39} &  \textbf{2.95} $\pm$ \footnotesize{0.24}&  \textbf{1.87} $\pm$ \footnotesize{0.14} & 6.18 $\pm$ \footnotesize{0.73}& 4.15 $\pm$ \footnotesize{0.39}&  3.16 $\pm$ \footnotesize{0.30}\\
LookAhead MPC-5           &  \underline{6.25} $\pm$ \footnotesize{0.51}&  3.90 $\pm$ \footnotesize{0.21} &  2.85 $\pm$ \footnotesize{0.38} &  \textbf{5.34} $\pm$ \footnotesize{0.09} &  \underline{2.97} $\pm$ \footnotesize{0.29}&  \underline{1.91} $\pm$ \footnotesize{0.18} & \underline{6.15} $\pm$ \footnotesize{0.47}&  \textbf{3.77} $\pm$ \footnotesize{0.67}&  \textbf{2.95} $\pm$ \footnotesize{0.47}\\
%MPC-H1 (vanilla) &  6.36 $\pm$ \footnotesize{0.26}&  4.35 $\pm$ \footnotesize{0.33} &  3.22 $\pm$ \footnotesize{0.29} &  5.80 $\pm$ \footnotesize{0.37} &  3.35 $\pm$ \footnotesize{0.13}&  2.38 $\pm$ \footnotesize{0.21} & 6.72 $\pm$ \footnotesize{0.37}&  4.77 $\pm$ \footnotesize{0.43}&  3.79 $\pm$ \footnotesize{0.45}\\
%MPC-H1           &  \underline{6.18} $\pm$ \footnotesize{0.70}&  \underline{3.55} $\pm$ \footnotesize{0.32} &  \textbf{2.55} $\pm$ \footnotesize{0.21} &  \underline{5.72} $\pm$ \footnotesize{0.53} &  \underline{3.05} $\pm$ \footnotesize{0.37}&  \underline{2.27} $\pm$ \footnotesize{0.30} & \textbf{6.12} $\pm$ \footnotesize{0.59}&  \textbf{3.74} $\pm$ \footnotesize{0.34}&  \textbf{2.87} $\pm$ \footnotesize{0.16}\\
\bottomrule
\end{tabular}
\end{adjustbox}

    \caption{Average normalized regret. We
report the best results in \textbf{bold} and \underline{underline} second best.}
    \label{tab:regret}
\end{table*}

We report in Tables~\ref{tab:rank} and \ref{tab:regret} the average rank and normalized regret, respectively, over 5 runs for 50 trials with three different seeds per run for all methods. Each transfer learning approach had access to the same training datasets and all models were initialized via the same three seeds to fit the initial surrogate. Overall, our LookAhead method outperforms the baselines across all meta-datasets. We use the horizon of $H3$ and $H5$ in this section, and study the contribution of LookAhead and the importance of planning in Section~\ref{sec:ablation}.

LookAhead MPC-3 and LookAhead MPC-5 demonstrate clear gains in the average rank with consistent performance against the baselines. Looking at the normalized regret, FSBO outperforms our model in some cases, e.g. Layout Md-v2, however, the associated average rank is still lower. This can be attributed to the fact that different tasks have heterogeneous response distributions. When there is a clear contradiction between the rank and the normalized regret, this signifies that the margin of difference of the normalized regret on the \textit{few} tasks where FSBO shows stronger performance, is higher than where it has lost. 

Comparing LookAhead MPC-3 and LookAhead MPC-5, we notice that a shorter horizon is sufficient to explore smaller search spaces. Albeit, early exploration in the form of longer horizons can still turn out to be more favorable, e.g in Regularization Md. 
%The average rank is a much faithful evaluation metrics. 
%This once again signifies that the average rank is a more faithful evaluation metric.
%Across the meta-datasets, there does not seem to be any clear runner-up approach. FSBO shows competitive behavior on Layout Md-v2 and Optimization Md-v2 but is outperformed, along with the rest of the baselines, by a simple GP on Regularization Md. This shows that some transfer learning solutions fail to capture the underlying similarities across tasks, and thus might not generalize well.  }

It is also worth mentioning that MetaBO, the model-free RL approach,  is the worst across all meta-datasets. This is attributed to the asymptotic degradation of the performance with the increasing number of trials, as pointed out in ~\cite{Wistuba2021_FSBO}. Contrary to MetaBO, which relies on an underlying surrogate to generate the state which is bound by a fixed grid, our approach is agnostic to the size of the grid and improves with more trials, leading to an improved state representation. %becomes richer.
\subsection{Ablation}
\label{sec:ablation}
\begin{figure}
    \centering
    \begin{subfigure}
    \centering
    \includegraphics[width=0.95\columnwidth]{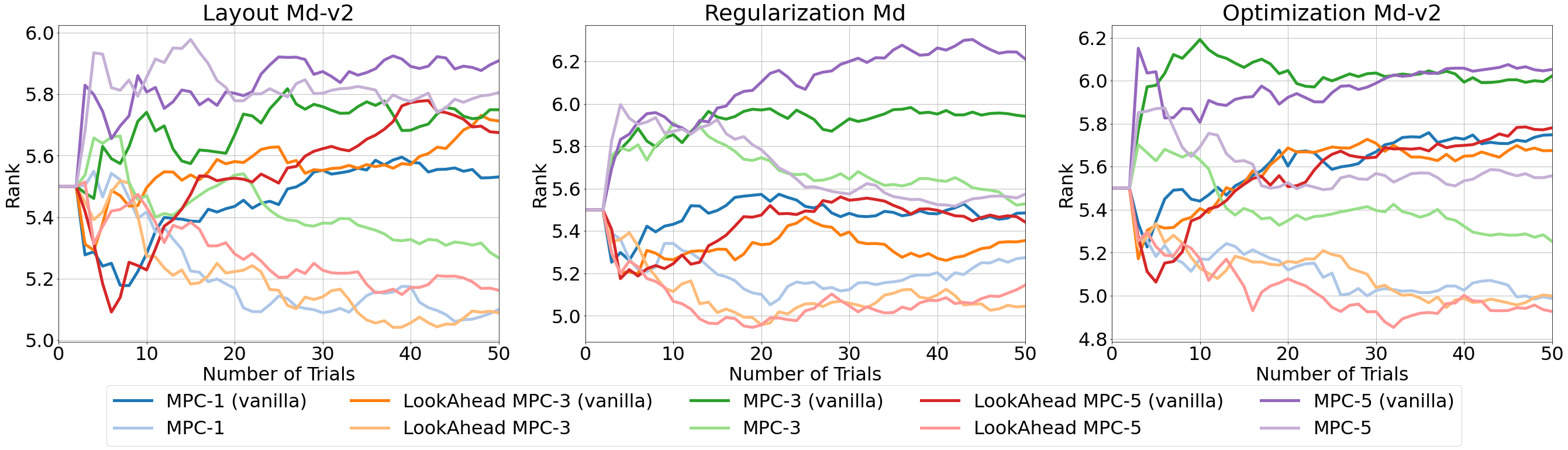}
    \end{subfigure}    
    \begin{subfigure}
    \centering
    \includegraphics[width=0.95\columnwidth]{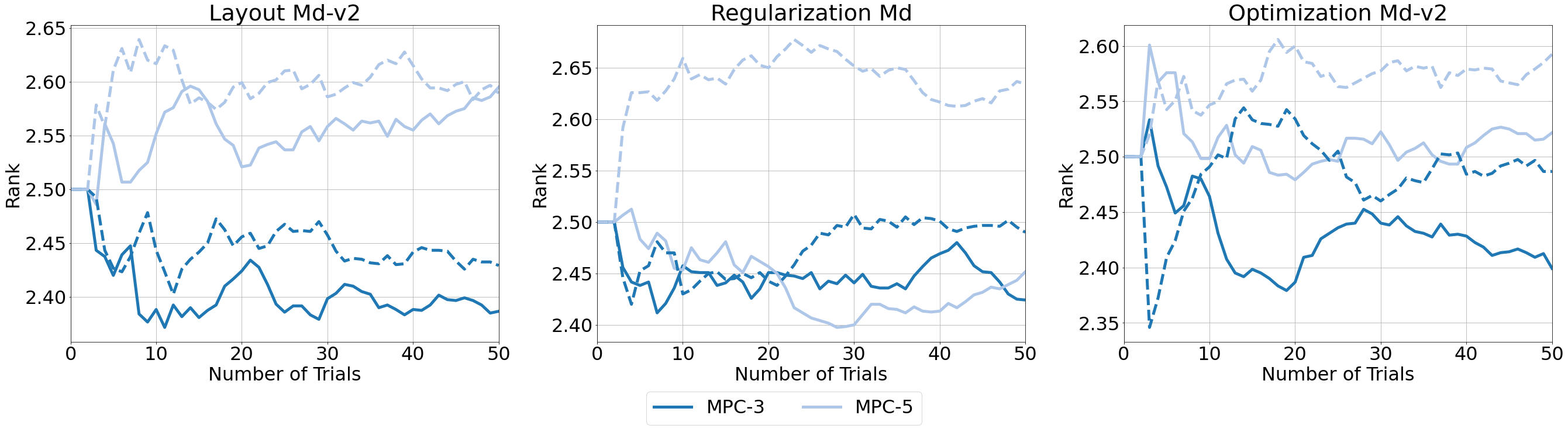}
    \caption{(top) Investigating the impact of planning and LookAhead across different horizons; (bottom) Increasing the number of sampled trajectories improves planning (solid lines$\rightarrow$ 1000 trajectories vs. dashed lines $\rightarrow$ 100 trajectories).}
    \label{fig:trajectoies}
    \end{subfigure}
\end{figure}

We further investigate the effect of LookAhead on MPC with different horizons and trajectory samples to evaluate the impact of planning under such conditions. We summarize the performance in Figures~\ref{fig:trajectoies} and share the following insights:
%\begin{enumerate}
%    \item 
i) Fine-tuning plays a crucial role in improving performance. Contrary to existing MbRL solutions that evaluate a policy’s performance on the environment which the transition model has been trained to estimate, we evaluate our policy on new environments, i.e. new tasks, with varying response surfaces. We notice that with a few gradient update steps, (LookAhead) MPC-\textbf{X} outperforms the vanilla variant.
    ii) Using LookAhead MPC is better than standard MPC. This reinforces the notion that via proper planning the evaluation of bad hyperparameter configurations can be avoided.
%    \item The performance steadily deteriorates as the rollout horizon increases. This is due to the compounded error and can be resolved with better transition models.
    iii) Increasing the number of sampled trajectories leads to a better outcome. This is to be expected, considering that for random shooting, we sample a fixed number of trajectories to evaluate. At any given trial $t$, the number of possible trajectories for a horizon $H$ on a grid of size $N$ is $\binom{N-t}{H}$, so for a horizon $H=1$, we default to maximization over the grid.%, and achieve better results.
%\end{enumerate}

\section{Conclusion}
In this paper, we present a novel solution for HPO within the context of MbRL. Specifically, we design a new surrogate as an ensemble of neural networks, which is initialized through meta-learning and thus can adapt quickly to new target datasets with few observations. We propose a novel acquisition function based on model predictive control that utilizes a simple lookahead strategy to select good hyperparameter candidates from simulated trajectories. To the best of our knowledge, we are the first to investigate the impact of planning on HPO and present an extensive ablation study that motivates research in that direction. 

\bibliography{iclr2022_conference}
\bibliographystyle{iclr2022_conference}
%We primarily investigate the effect of domain adaptation on our learned transition model. Contrary to existing MbRL solutions that evaluate a policy’s performance on the environment which the transition model has been trained to estimate, we evaluate our policy on new environments, i.e. new tasks, with varying response surfaces. We notice that with a few gradient update steps, MPC-1 consistently outperforms its variant, MPC-1 (vanilla). Interestingly, in Layout Md-v2, MPC-1 (vanilla) is even competitive with the state-of-the-art transfer learning solutions without any adaptation steps, reflecting strong  generalization behavior  as  the policy  navigates  the  joint  approximate  transition model of the training datasets.
\clearpage
\appendix
\section{Algorithm}
Meta-learning has found resounding success in the research community as an initialization scheme, which allows for fast adaption to new domains.
We want to emphasize that during meta-training the meta-test datasets are not observed and in that way remain strictly held out.
% want to put this in the appendix?

%%% ALGORITHM %%%
\begin{algorithm}\caption{Model-based Reinforcement Learning}\label{alg:pseudocode}
\begin{algorithmic}[1]
\STATE{\algorithmicrequire~ training dataset $\mathcal{E}^\text{train}$; parameters $\theta$; learning rate $\eta$; inner update steps $v$; meta-batch size $n$}
\STATE{\textbf{while} not converged \textbf{do}}
\STATE{~~~~~~~$t\sim\text{Unif}\left([T_\text{min},T_\text{max}]\right)$}
\STATE{~~~~~~~$D_1,\dots,D_n\sim$~Unif$\left([1,\dots, N]\right)$}
\STATE{~~~~~~~\textbf{for} $i=1$~to $n$ \textbf{do}}
\STATE{~~~~~~~~~~~~~~$\left(s_t,a_t,\ell^{(D_i)}(a_t)\right)\sim\text{Unif}\left(\mathcal{E}^\text{train} \mid D_i, t\right)$}
\STATE{~~~~~~~~~~~~~~$\theta_i\leftarrow\theta$}
\STATE{~~~~~~~~~~~~~~\textbf{for} $j=1$~to~$v$ \textbf{do}}
\STATE{~~~~~~~~~~~~~~~~~~~~~$\theta_i\leftarrow\theta_i + \eta\nabla_\theta\left(p_\theta(s_t\mid s_t,a_t)\right)$}
\STATE{~~~~~~~Update $\theta \leftarrow\theta + \eta\frac{1}{n}\sum_{i=1}^{n}\left(\theta_i-\theta\right)$}
\STATE{\textbf{return} $\theta$}
\end{algorithmic}
\end{algorithm}
\section{Meta-dataset}
\label{appendix:metadataset}
A meta-dataset is a collection of hyperparameters, typically defined on a \textit{discretized} grid~\citep{Schilling2016_PoG, Jomaa2019_HYPRL}, associated with a model under investigation, that have been evaluated offline by training the model with the mentioned hyperparameters on numerous \textit{primary} datasets, and reporting some observed evaluation metric, e.g. validation loss. 

We evaluate our approach on three hyperparameter search spaces for feed-forward neural networks \citep{Jomaa2021_DMFBS}, that includes 120 UCI classification datasets~\citep{Asuncion2007_UCI}. 
\cite{Jomaa2021_DMFBS} propose a pruning strategy that eliminates \textit{redundant} configurations. Specifically, they drop certain hyperparameter combinations, e.g. non-$\square$ layouts with 1 hidden layer, which we see here as unnecessary. Therefore we do not prune \textbf{Layout Md} and \textbf{Optimization Md} and refer to the meta-datasets as \textbf{Layout Md-v2}, \textbf{Regularization Md}, and \textbf{Optimization Md-v2} where each meta-dataset includes the Cartesian product of the individual hyperparameters with a total of 324, 288, and 432 unique configurations, respectively.  %Figure~\ref{fig:kde} highlights the significant distribution shift of selected datasets across the meta-datasets. 
\begin{comment}
\begin{figure}
    \centering
    \includegraphics[width=\columnwidth]{distribution-shift.png}
    \caption{Kenel density estimation of selected datasets across the three meta-datasets}
    \label{fig:kde}
\end{figure}
\end{comment}
 \begin{table}[ht]
 \caption{Hyperparameter search space for the meta-datasets.}
 \centering
 \centering
  \begin{tabular}{llll}
 \toprule
 Hyperparameter &  Layout Md-v2 & Regularization Md &   Optimization Md-v2\\
 \midrule
 Activation   & ReLU, SeLU    &  ReLU, SeLU, LeakyReLU     &  ReLU, SeLU, LeakyReLU \\
 Neurons   &  $4,8,16,32$    &   $4,8,16,32$    &   $4,8,16$ \\
 Layers   & $1,3,5,7$    &  $1,3,5,7$     &  $3,5,7$ \\
 Layout  &  $\square$,$\lhd$,$\rhd$,$\diamond$,$\bigtriangleup$   &   $\square$    &  $\lhd$,$\rhd$,$\diamond$,$\bigtriangleup$ \\
 Dropout&$0,0.5$ & 0, 0.2, 0.5 & $0$ \\
 Normalization& False &  False, True & False \\
 Optimizer& ADAM &  ADAM  & ADAM, RMSProp, GD \\
 \bottomrule
 \end{tabular}
 \label{grid}
 \end{table}
 
The aforementioned hyperparameters are encoded as follows:
 
 \begin{table}[ht]
 \caption{Hyperparameter Encoding}
 \centering
 \begin{tabular}{ll}
 \toprule
 Hyperparameter & Encoding \\
 \midrule
 Activation   & One-hot encoding\\
 Neurons   &  Scalar\\
 Layers   & Scalar \\
 Layout  &  One-hot encoding\\
 %Layout~\cite{jomaa2019dataset2vec} & squ & asc,des,diamond,enc & - & grow,decay,diamond,enc w/o squ\\
 Dropout& Scalar\\
 Normalization& Scalar\\
 Optimizer& One-hot encoding\\
 \bottomrule
 \end{tabular}
 \end{table}
 
\section{Baselines}
\label{appendix:baselines}
\begin{enumerate}
    \item Random sampling~\citep{Bergstra2012_Random},
    \item GP~\citep{Rasmussen2003_GP} is a hyperparameter tuning strategy that relies on a Gaussian process as a surrogate model with squared exponential kernels (Matern 5/2 kernel) with automatic relevance determination,
    \item SMAC~\citep{Hutter2011_SMAC} utilizes random forests instead of Gaussian processes to represent the surrogate model,
    \item BOHAMIANN~\citep{Springenberg2016_Bohamiann} is based on Bayesian neural networks that are trained via stochastic gradient Hamiltonian Monte Carlo,
    \item TST-R~\citep{Wistuba2016_TSTR} is an ensemble approach where the Gaussian process surrogate of the target task is weighted with surrogates of the training datasets based on the ranking similarity of the evaluated hyperparameters,
 %   \item RGPE~\citep{Feurer2018_RGPE} is another ensemble approach, similar to TST-R, which estimates the weights by optimizing a ranking loss between the surrogates of the training datasets and that of the target task,
    \item ABLR~\citep{Perrone2018_ABLR} is a multi-task Bayesian linear regression approach that optimizes a shared feature extractor across the training datasets as an initialization strategy for the target task,
    \item CTS~\citep{Salinas2020_CTS} trains a Gaussian Copula process~\citep{Wilson2010_Copula} jointly over the training datasets mapped to a shared output distribution using quantile-transformations. Hyperparameter candidates are selected via Thompson Sampling,
    \item FSBO~\citep{Wistuba2021_FSBO} uses deep Kernel Gaussian processes~\citep{Gordon2015_DKL} to estimate the response of the target dataset. The parameters are initialized via meta-learning the joint response surface over the training datasets.
\end{enumerate}

\end{document}